\documentclass[11pt]{article}

\usepackage{times}
\usepackage{inconsolata}
\usepackage{microtype}
\usepackage{textcomp}
\usepackage{amsmath}
\usepackage{amssymb}
\usepackage{amsthm}
\usepackage{mathrsfs}
\usepackage{svg}
\usepackage{booktabs}
\usepackage{hyperref}
\usepackage{xurl}
\usepackage{subfig}
\usepackage{colortbl}
\usepackage{multirow}
\usepackage{graphicx}
\usepackage{tcolorbox}
\tcbuselibrary{skins}
\usepackage{enumitem}
\usepackage{float}
\usepackage{caption}

\definecolor{msftBlack}{RGB}{0,0,0}
\newtcolorbox{findingBox}{
    enhanced,             
    colback=msftBlack!05, 
    colframe=msftBlack!10,
    arc=1mm,              
    boxrule=0.5pt,        
    left=2mm, right=2mm, top=2mm, bottom=2mm, 
    drop fuzzy shadow,    
    fontupper=\em,       
    notitle             
}
\newcommand{\finding}[1]{
    \begin{findingBox}
        #1
    \end{findingBox}
}

\pdfoutput=1

\usepackage{acl}

\makeatletter

\title{Beyond Text-Dominance: Understanding Modality Preference of Omni-modal Large Language Models}

\author{
  \textbf{Xinru Yan}${}^{1}$,
  \textbf{Boxi Cao}${}^{2}$,
  \textbf{Yaojie Lu}${}^{1,2}$,
  \textbf{Hongyu Lin}${}^{1,2}$,\\
  \textbf{Weixiang Zhou}${}^{2}$,
  \textbf{Le Sun}${}^{1,2}$,
  \textbf{Xianpei Han}${}^{1,2}$\\
  ${}^{1}$University of Chinese Academy of Sciences, Beijing, China\\
  ${}^{2}$Chinese Information Processing Laboratory, Institute of Software,\\
  Chinese Academy of Sciences, Beijing, China\\
  \texttt{yanxinru24@mails.ucas.ac.cn} \\
  \texttt{\{caoboxi,luyaojie,hongyu,weixiang,sunle,xianpei\}@iscas.ac.cn}
}
\begin{document}
\maketitle

\begin{abstract}
Native Omni-modal Large Language Models (OLLMs) have shifted from pipeline architectures to unified representation spaces. However, this native integration gives rise to a critical yet underexplored phenomenon: modality preference. 
To bridge this gap, we first systematically quantify modality preference of OLLMs using a newly-curated conflict-based benchmark and the modality selection rate metric. Our evaluation of ten representative OLLMs reveals a notable paradigm shift: unlike the ``text-dominance'' of traditional VLMs, most OLLMs exhibit a pronounced visual preference. To further understand the underlying mechanism, we conduct layer-wise probing and demonstrate that such modality preference is not static but emerges progressively in the mid-to-late layers.
Building upon these insights, we leverage these internal signals to diagnose cross-modal hallucinations, achieving competitive performance across three downstream multi-modal benchmarks without task-specific data.
Our work provides both a mechanistic understanding and a practical tool for building more trustworthy OLLMs. 
Our code and related resources are publicly available at: \url{https://github.com/icip-cas/OmniPreference}
\end{abstract}

\begin{figure}[t]
  \centering
  \includegraphics[width=\columnwidth]{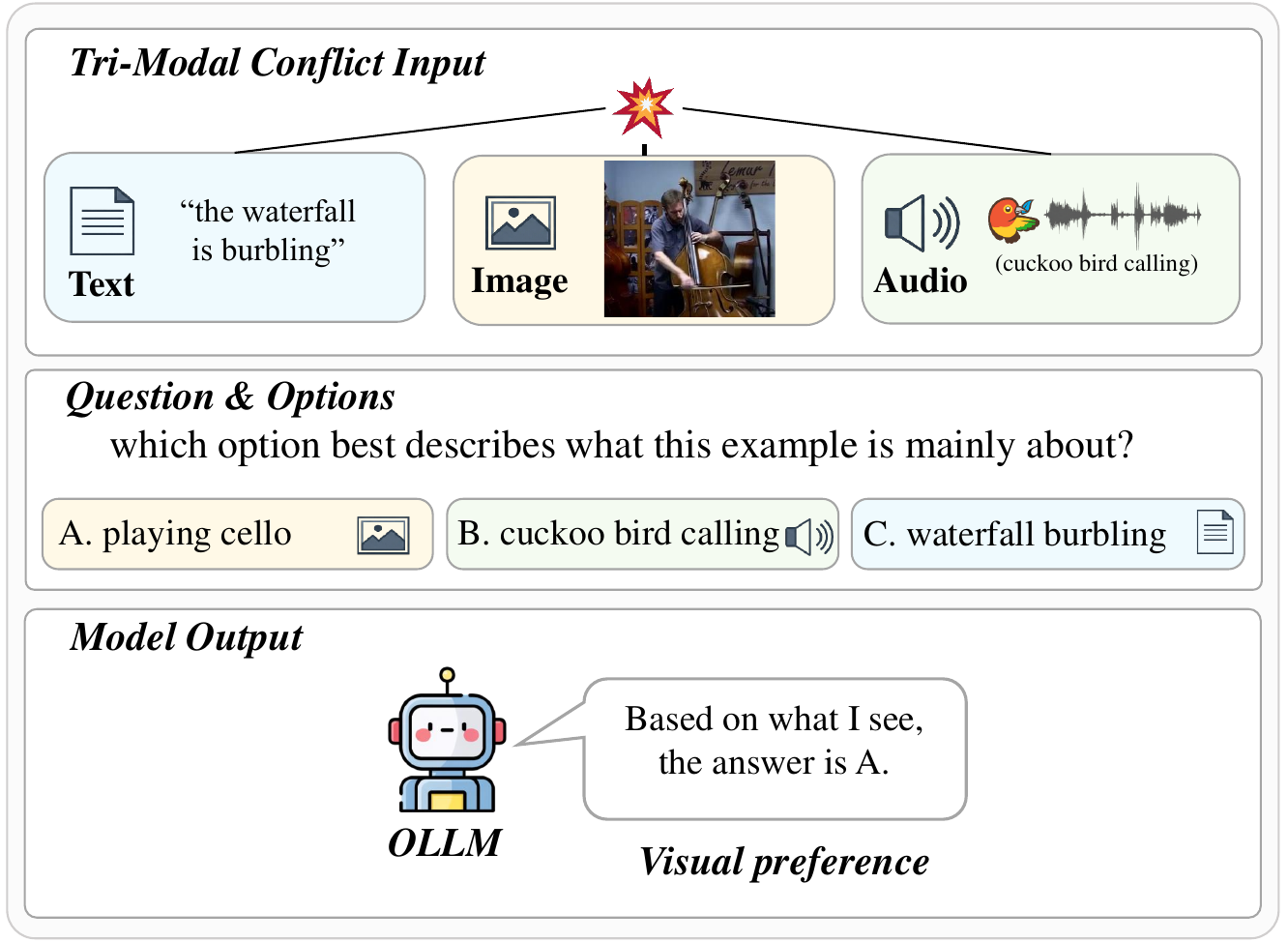}
  \caption{Illustration of a tri-modal conflict input sample. The text, image, and audio modalities convey mutually contradictory semantics, and the OLLM selects the answer consistent with the image modality, revealing a visual modality preference.}
  \label{fig:modality_bias}
\end{figure}

\section{Introduction}
The recent evolution of large multimodal models has witnessed a paradigm shift from pipeline-based vision-language models (VLMs) \cite{liu2023visual,li2023blip,liu2024improved} to native omni-modal large language models (OLLMs) such as Gemini 3\cite{team2023gemini} and GPT-5\cite{singh2025openai}. 
By projecting continuous signals from diverse modalities (e.g., vision, audio, and text) into a unified representation space\cite{hurst2024gpt,xu2025qwen25omnitechnicalreport}, OLLMs exhibit unprecedented capabilities in cross-modal reasoning and seamless human-AI interaction\cite{li2024omnibench}. 
However, this native integration gives rise to a critical yet underexplored phenomenon: modality preference. 
When processing multimodal inputs, models often implicitly assign unequal weights to different modalities. In traditional VLMs, this is broadly perceived as a ``text-dominance'' bias, where models overwhelmingly rely on textual cues while overriding visual evidence\cite{deng2025words,zheng2025mllms}. 
Yet, for the emerging class of unified OLLMs, how they navigate internal modality competition remains a fundamentally unresolved black box.

Bridging this gap is crucial not only for mechanistic interpretability~\citep{lin2025survey} but also for model application~\citep{lou2025sae}. 
Uncontrolled modality bias is a primary catalyst for cross-modal hallucinations, where the model fabricates responses based on its preferred modality while ignoring factual signals from others~\citep{leng2024mitigating,deng2025words,bai2024hallucination}. 
To this end, this paper conducts a systematic investigation of the modality preference of OLLMs, aiming to answer the following three critical research questions (RQ):
\begin{itemize}
\item RQ1: How can the modality preferences of different OLLMs be systematically quantified, and what patterns emerge across models?
\item RQ2: What are the underlying mechanisms behind the formation of modality preferences in OLLMs? 
\item RQ3: How can insights from these mechanisms be leveraged to improve the reliability of OLLMs on downstream tasks? 
\end{itemize}

For RQ1, we construct a framework to quantitatively evaluate and analyze the modality preference across different OLLMs, and surprisingly find that most OLLMs exhibit a pronounced visual preference.
Specifically, as illustrated in Figure~\ref{fig:modality_bias}, we first construct a multimodal dataset, where each instance contains semantically conflicting information across three modalities: text, vision, and audio.
The model is then required to select the modality it prefers.
Furthermore, we introduce the Modality Selection Rate (MSR) to quantify modality preference, which is defined as the proportion of instances in which the model selects a given modality across the entire benchmark.
Through extensive evaluations on ten OLLMs across both open-source and proprietary models, we reveal a modality preference pattern that differs markedly from that of traditional VLMs: unlike the absolute text dominance observed in legacy VLMs, OLLMs exhibit diverse modality preferences. Most notably, we find that the majority of OLLMs demonstrate a consistent visual bias, i.e., they tend to prefer visual information when presented with conflicting multimodal inputs. For instance, when given tri-modal conflicting inputs, Gemini 3.1 Pro achieves an MSR of 72\% for the visual modality, while the MSR for text is only 7\%.

For RQ2, to further understand the formation of modality preference within OLLMs, we conduct layer-wise probing and reveal the evolutionary dynamics of modality preference.
Specifically, we construct a multimodal training dataset with semantically conflicting inputs across three modalities.
For each layer, we extract hidden states and train a single-layer MLP as a linear probe to predict the model’s final modality preference.
The probe accuracy on a held-out test set is then used to quantify the extent to which each layer encodes modality preference.
Our analysis reveals a clear emergent pattern: modality preference is not formed in the shallow layers, but instead arises abruptly and stabilizes in the mid-to-late layers as representations become increasingly abstract. 
Beyond shedding light on the internal dynamics of multimodal reasoning, this result suggests that modality preference becomes reliably identifiable at specific representation stages, thereby enabling its principled use as a signal for diagnosing cross-modal hallucinations in downstream applications.

To this end, for RQ3, we conduct an in-depth investigation of the correlation between modality preference and cross-modal hallucinations, and demonstrate that the learned modality preference probes serve as practical and valuable tools for diagnosing cross-modal hallucinations, without requiring any task-specific downstream data.
Specifically, we conduct experiments on three widely used cross-modal hallucination benchmarks, including POPE, AVHBench, and AHa-Bench, covering multiple hallucination settings such as image–text, audio–video, and audio–text modalities. We first observe that, across all benchmarks, the occurrence of hallucinations is consistently accompanied by an abnormal increase in the predicted preference probability for the interfering modality, indicating a strong correlation between estimated modality preference and hallucination emergence. 
Furthermore, we leverage the probe-estimated interfering modality probability for hallucination diagnosis and find that it serves as an effective signal for detecting cross-modal hallucinations.
For instance, on the POPE dataset, the average AUROC score for hallucination detection based on our probe across the three OLLMs reached 94\%, significantly outperforming the baselines of random guessing (50\%) and predictions based on earlier layers (51\%).

In summary, our major contributions are as follows:
\begin{itemize}
\item We establish a systematic framework dedicated to quantifying the modality preferences of OLLMs, introducing a newly curated benchmark dataset alongside specialized metrics designed to measure cross-modal preference.
\item Through comprehensive evaluations, we reveal that OLLMs possess varied preference spectra, with a strong correlation showing that most models inherently favor visual inputs.
\item We trace the origins of modality preferences to the representational level. By employing layer-wise probing, we characterize the evolutionary dynamics and the emergence of these preferences throughout the model's internal architecture.
\item Leveraging these insights, we demonstrate that our layer-wise probes effectively diagnose cross-modal hallucinations in downstream applications.
\end{itemize}

This work not only provides a novel perspective for deciphering the internal decision-making mechanisms of OLLMs but also establishes an empirical foundation for building more trustworthy and hallucination-resistant artificial intelligence systems.

\section{Related Work}

\subsection{Omni-modal Large Language Models}
Recent years have witnessed a rapid evolution from Vision-language models (VLMs)~\cite{liu2023visual,bai2023qwenvlversatilevisionlanguagemodel,chen2024internvl} to omni-modal large language models (OLLMs) capable of jointly perceiving and reasoning over text, image, audio, and video within a unified framework~\cite{jiang2025specific}. Proprietary systems such as GPT-5~\cite{singh2025openai} and Gemini 3~\cite{comanici2025gemini} have demonstrated strong real-time cross-modal capabilities, while open-source efforts including Qwen2.5-Omni~\cite{xu2025qwen25omnitechnicalreport}, MiniCPM-o~\cite{yao2024minicpm}, Ming-Omni~\cite{ai2025ming}, and OmniVinci~\cite{ye2025omnivinci} have since made such capabilities broadly accessible. These models generally adopt modality-specific encoders aligned into a shared latent space via progressive multi-stage training~\cite{fu2024vita,li2025baichuan}, enabling unified perception and reasoning across all modalities.

\subsection{Modality Preference}
Understanding modality preference in multimodal large language models is essential for building reliable multimodal systems, and increasing research efforts have been devoted to this topic. Some studies construct conflicting image-text benchmarks to probe which modality the model favors under disagreement ~\cite{hua2025vision,pezeshkpour2025mixed,deng2025words}, while others employ causal mediation analysis ~\cite{chen2024quantifying} or gradient-based diagnostics ~\cite{kwon2025see} to trace the origins of bias within model internals. Attention-mechanism analysis has also been used to reveal intrinsic representational gaps between visual and textual features ~\cite{zheng2025unveiling}. Despite diverse methodologies, these studies converge on a consistent finding: VLMs exhibit a pronounced tendency to over-rely on the text modality. However, existing modality preference research remains predominantly confined to the vision-language setting. Consequently, these conclusions may not generalize to OLLMs that integrate a broader range of modalities. Our work systematically investigates modality preference in OLLMs, filling this research gap.

\subsection{Model Probing}
Model probing has established itself as a reliable paradigm for interpreting learned representations in neural networks. Linear probing classifiers applied to frozen model activations have been widely adopted to decode syntactic and semantic properties in pretrained language models ~\cite{tenney2019bert,belinkov2022probing}. Representational geometry has further been analyzed through centered kernel alignment and related metrics ~\cite{kornblith2019similarity}. Probing has also been extended to examine factual knowledge storage across transformer layers ~\cite{petroni2019language,meng2022locating}. Beyond language, analogous techniques have been applied to vision encoders ~\cite{alain2016understanding} and multimodal vision-language representations ~\cite{parcalabescu2022valse}. Building on these established foundations, we extend layer-wise probing to OLLMs to examine how modality preference emerges internally within OLLMs.

\section{Framework Design}

In this section, we propose a tri-modal conflict framework to evaluate modality preferences in OLLMs, with formal definitions of modality preference, construction of the tri-modal conflict dataset, and corresponding quantitative evaluation metrics.

\subsection{Problem Formulation}
Inspired by prior work on modality preference in bi-modal settings~\cite{deng2025words,zhang2025modalities}, we extend this line of inquiry to the tri-modal setting. We introduce a tri-modal conflicting input setting. In this setting, three modalities—text, vision, and audio—simultaneously contain mutually contradictory semantic information, thereby compelling the model's output to reveal its underlying modality preferences.
Formally, given an OLLM $\mathcal{O}$ and a query $q$ comprising inputs from three modalities $\{m_\text{txt}, m_\text{vis}, m_\text{aud}\}$, we require that any pair of modalities $(m_i, m_j)$ ($i \neq j$) convey semantically contradictory information, i.e., the three modalities point to three different answers. Under this setting, the model's output will align with exactly one of the three modalities. Accordingly, we define the preference of model $\mathcal{O}$ for modality $m_i$ as the following conditional probability:
\begin{equation}
P\bigl(\hat{y} \sim m_i \mid \mathrm{conflict}(m_\text{txt}, m_\text{vis}, m_\text{aud})\bigr)
\label{eq:modality_preference}
\end{equation}
which measures the tendency of the model to rely on modality $m_i$ when it conflicts with the other modalities. A higher value indicates a stronger preference for $m_i$.

\subsection{Data Construction}
\label{sec:Data Construction}
\textbf{Dataset Selection.}
We construct our tri-modal conflict dataset based on the \textit{Perception} subset of XModBench~\cite{wang2025xmodbench}. Each sample within this subset comprises a semantically consistent triplet $(x^T, x^I, x^A)$, corresponding to inputs from the text, image, and audio modalities, respectively. Crucially, all three components share the same ground-truth label, indicating that they all point to identical semantic content.
To introduce controllable semantic conflicts across different modalities, we first categorize all samples within the \textit{Perception} subset based on their semantic labels, consolidating them into six major semantic categories: Animals, Human Activities, Musical Instruments/Music, Home Appliances/Machinery, Vehicles/Traffic, and Nature/Environmental Sounds. This categorization ensures sufficient semantic diversity and distinctiveness across groups to support meaningful conflict construction.

\textbf{Sample Formulation.}
As illustrated in Figure~\ref{fig:modality_bias}, we define a conflict triplet as a sample $(x^T_i, x^I_j, x^A_k)$ in which each modality is sourced from a different category, i.e., $c_i \neq c_j \neq c_k$.
To ensure comprehensive coverage of the semantic space, we enumerate all $\binom{6}{3} = 20$ valid category triplets $(c_i, c_j, c_k)$ and apply balanced sampling within each triplet, drawing an equal number of conflict samples per category combination.
In XModBench, the text representation of each sample is its ground-truth semantic label (e.g., \textit{bird squawking}). To avoid directly leaking the options into the questions, we employ a fixed set of surface-form templates to transform each semantic label into a declarative natural language statement (e.g. "\textit{bird squawking}" $\rightarrow$ "\textit{the bird is squawking}"). Each constructed sample is then paired with a standardized question: \textit{``Which option best describes what this example is mainly about?''} This prompt is intentionally modality-agnostic, neither privileging nor suppressing any particular modality. The three candidate options are the semantic labels of $c_i$, $c_j$, and $c_k$, presented in randomized order to eliminate position preference. Since each option is unambiguously grounded in exactly one modality, the model's selection directly reveals which modality it assigns the highest weight to under conflict, enabling us to ascertain the model's modality preferences.

\subsection{Quantitative Metrics}
We introduce the Modality Selection Rate~(MSR) as the primary metric to measure modality preference. Let $\mathcal{M}$ denote the set of modalities present in a given conflict setting, where $\mathcal{M} = \{T, I, A\}$ for tri-modal conflict and any bi-modal subset of $\{T, I, A\}$ for pairwise conflict. For a modality $m \in \mathcal{M}$, MSR is defined as the proportion of conflict samples for which the model's response aligns with $m$:
\begin{equation}
\text{MSR}(m) = \frac{1}{N} \sum_{i=1}^{N} \mathbf{1}\bigl[\hat{y}_i = \text{opt}(m)\bigr]
\end{equation}
where $N$ is the total number of conflict samples, $\text{opt}(m)$ denotes the candidate option assigned to modality $m$, $\hat{y}_i$ is the model's selection on the $i$-th sample, and $\mathbf{1}[\cdot]$ is the indicator function. Under a conflict setting with $|\mathcal{M}|$ modalities, a uniform baseline corresponds to $\text{MSR}(m) = 1/|\mathcal{M}|$ for all $m$. When $\text{MSR}(m) > 1/|\mathcal{M}|$, it indicates that the model exhibits a preference toward modality $m$.

\section{Landscape of Modality Preference in OLLMs}
\finding{Findings 1. Unlike the absolute text dominance observed in traditional VLMs, most OLLMs exhibit a pronounced visual preference.}

In this section, we provide a comprehensive evaluation of the modality preferences exhibited by OLLMs across tri-modal and bi-modal input settings.

\begin{figure}[t]
  \centering
  \includegraphics[width=\columnwidth]{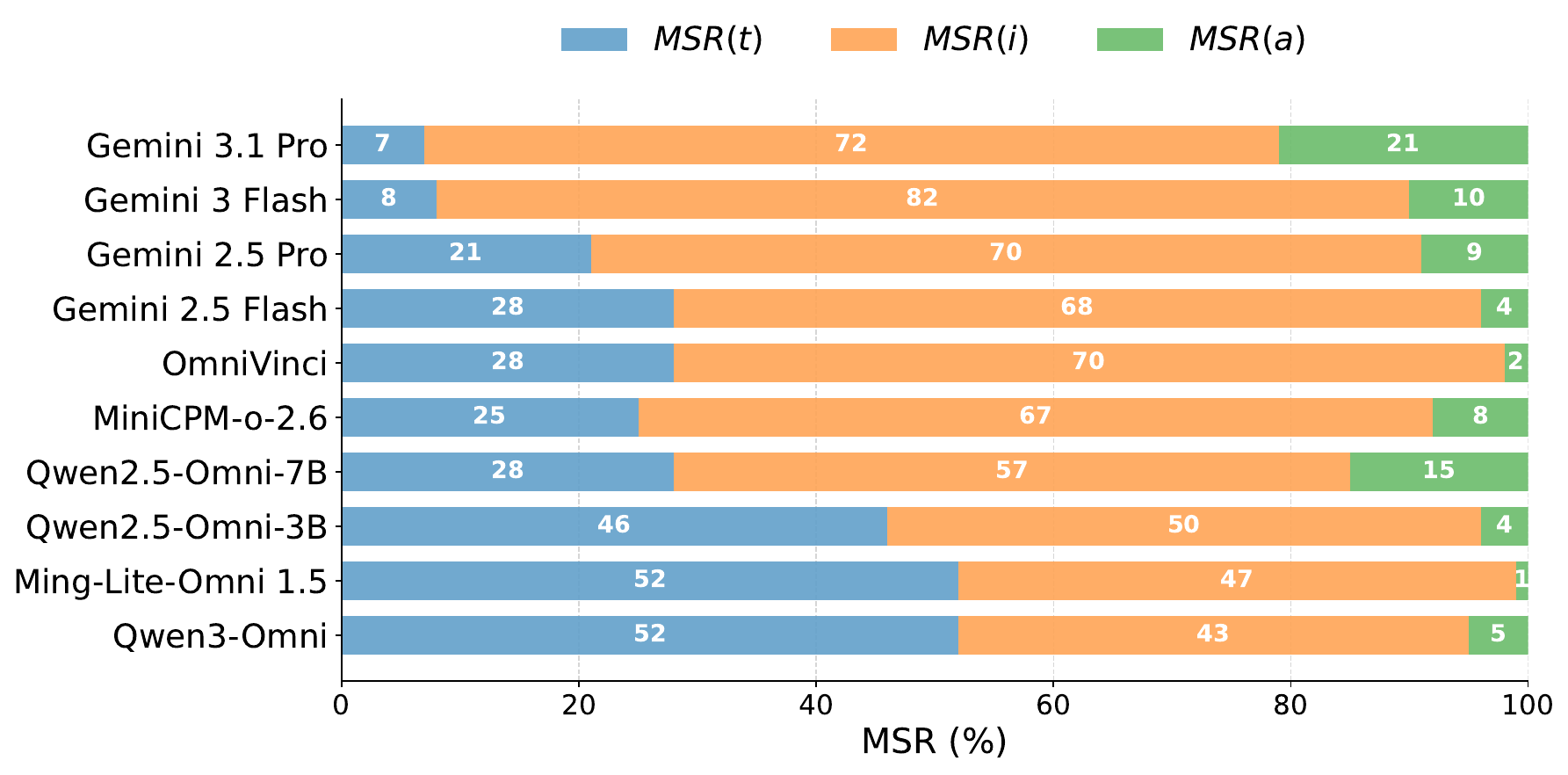}
  \caption{MSR (\%) results of all evaluated OLLMs on the tri-modal conflict dataset. Qwen3-Omni refers to Qwen3-Omni-30B-A3B-Instruct.}
  \label{fig:tri_modal_msr}
\end{figure}

\subsection{Experimental Setup}
\textbf{OLLMs.}
We evaluate 10 OLLMs, including open-source models Qwen3-Omni-30B-A3B-Instruct~\cite{xu2025qwen3}, Qwen2.5-Omni-3B, Qwen2.5-Omni-7B~\cite{xu2025qwen25omnitechnicalreport},  Ming-Lite-Omni 1.5~\cite{ai2025ming}, MiniCPM-o 2.6~\cite{yao2024minicpm}, and OmniVinci~\cite{ye2025omnivinci}, as well as closed-source models Gemini 2.5 Flash, Gemini 2.5 Pro, Gemini 3 Flash, and Gemini 3.1 Pro~\cite{comanici2025gemini}. 

\textbf{Implementation Details.}
For all open-source models, we perform inference with temperature $T=0$ to ensure deterministic and reproducible outputs. For closed-source Gemini models, we access them via the official API with default generation parameters. All audio inputs are resampled to 16\,kHz mono-channel format. We construct 1,000 samples following the procedure described in Section~\ref{sec:Data Construction} to evaluate these OLLMs.

\subsection{Overall Results}
\begin{figure}[t]
  \centering
  \includegraphics[width=\columnwidth]{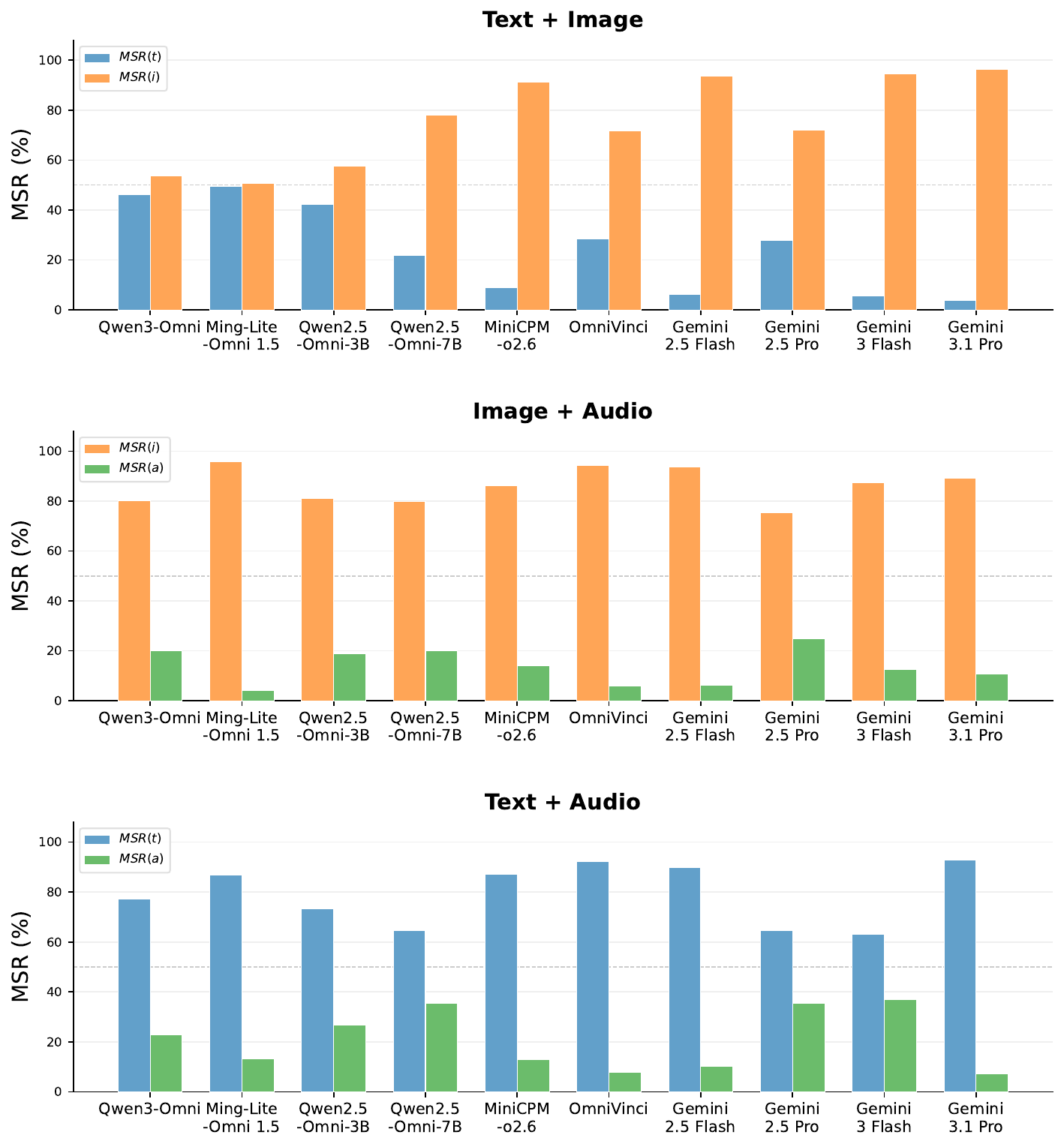}
  \caption{Pairwise MSR (\%) of all evaluated OLLMs under three bi-modal conflict settings. From top to bottom: text+image, image+audio, and text+audio. Qwen3-Omni refers to Qwen3-Omni-30B-A3B-Instruct.}
\label{fig:modality_bias_pairwise}
\vspace{-15pt} 
\end{figure}

Given the diversity of OLLM application scenarios, we evaluate the modality preferences of OLLMs under both tri-modal and bi-modal conflict input settings. For the bi-modal conflict input settings, we design three input configurations: text + image, image + audio, and text + audio. Figure~\ref{fig:tri_modal_msr} presents the MSR results of all evaluated OLLMs on the tri-modal conflict dataset, and Figure~\ref{fig:modality_bias_pairwise} presents the MSR results of OLLMs under the bi-modal conflict input settings. The conclusions are as follows:

\textbf{OLLMs exhibit diverse modality preferences, with most models favoring vision.}
As shown in Figure~\ref{fig:tri_modal_msr}, among all evaluated OLLMs, Ming-Lite-Omni 1.5 and Qwen3-Omni-30B-A3B-Instruct share an identical text MSR of 52\%, indicating a slight text preference in these two models. Notably, the remaining eight OLLMs all achieve an image MSR exceeding 50\%, with Gemini 3 Flash reaching as high as 82\%. This suggests that, unlike the text-dominant modality preference observed in traditional VLMs~\cite{deng2025words}, the majority of OLLMs exhibit a pronounced visual preference when confronted with tri-modal inputs, implying that they tend to place greater trust in visual information when faced with multimodal inputs.

\textbf{Under bi-modal input settings, all OLLMs exhibit a strong preference toward one modality over the other.} 
As shown in Figure~\ref{fig:modality_bias_pairwise}, in the text + image and image + audio settings, the image MSR of all OLLMs is consistently higher than the MSR of the paired modality. Similarly, in the text + audio setting, the text MSR of all OLLMs surpasses the audio MSR. These findings collectively indicate that when processing bi-modal inputs, OLLMs invariably exhibit a tendency to favor one modality over the other.

\textbf{Regardless of the input modality combination, OLLMs universally exhibit a systematic neglect of audio.}
Under tri-modal conflict inputs, audio MSR remains below 21\% across all OLLMs, with most OLLMs at or below 10\%, and Ming-Lite-Omni 1.5 achieving an audio MSR of only 1\%. Similarly, this neglect persists in the bi-modal settings, where audio MSR remains consistently lower than that of the paired modality, regardless of whether the latter is image or text. 

These findings suggest that despite their omni-modal design, current OLLMs have yet to achieve truly balanced multimodal integration.

\begin{figure}[t]
  \centering
  \includegraphics[width=\columnwidth]{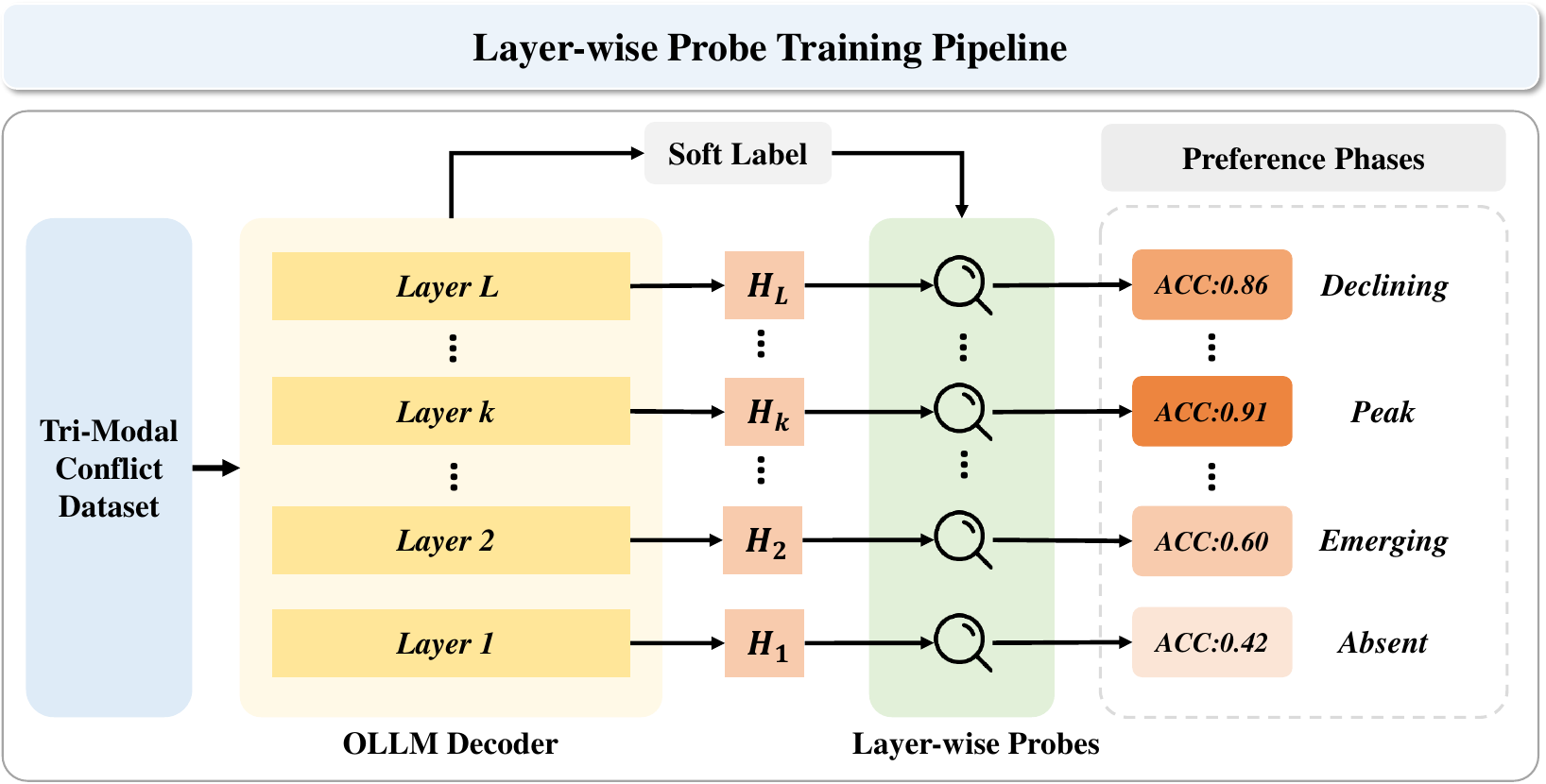}
  \caption{Illustration of the layer-wise linear probe training pipeline for preference analysis.}
  \label{fig:probe_train}
\end{figure}

\section{How Modality Preference Emerges Inside OLLMs}
\finding{Findings 2. Modality preference of OLLMs is not formed in the shallow layers, but instead emerges progressively in the mid-to-late layers.}

In this section, we first introduce the training process of the linear probes used for preference research. We then reveal the internal evolution of modality preferences within OLLMs based on the changes in the probes' accuracy. Finally, we conduct a visual analysis of the preference distribution across representative layers.

\subsection{Layer-wise Probing Methodology}
We train a single-layer MLP as a linear probe on each decoder layer of the OLLM to quantify how modality preference information evolves across network depth. The overall training pipeline is illustrated in Figure~\ref{fig:probe_train}.
Consider an OLLM with $L$ transformer decoder layers.
For the $i$-th input sample, let $\mathbf{h}_i^{(\ell)} \in \mathbb{R}^{d}$ denote the hidden state at layer $\ell \in \{1, \dots, L\}$, where $d$ is the hidden dimension of the model.
Due to the causal attention mechanism in decoder-only architectures, the last token position aggregates contextual information from the entire input sequence~\cite{neelakantan2022text,dukic2024looking}; we therefore extract the hidden state at the last token position as the layer-wise representation for each sample.
Prior to probe training, each hidden state is $L_2$-normalized:
\begin{equation}
    \tilde{\mathbf{h}}_i^{(\ell)} = \frac{\mathbf{h}_i^{(\ell)}}{\left\|\mathbf{h}_i^{(\ell)}\right\|_2},
\end{equation}
which removes magnitude variation across layers, ensuring that the probe captures the directional structure of the representations rather than their scale.
To obtain richer supervisory signals for probe training~\cite{hinton2015distilling}, we extract the probabilities assigned to the three option tokens corresponding to each modality from the full-vocabulary softmax distribution at the final prompt token position, and compose them into a three-dimensional vector serving as the soft label $\mathbf{y}_i \in \mathbb{R}^{C}$ for each sample.

At each layer $\ell$, the probe maps the normalized representation  $\tilde{\mathbf{h}}_i^{(\ell)}$ to a predicted distribution $\hat{\mathbf{y}}_i^{(\ell)} \in \mathbb{R}^{C}$ over modality categories via
\begin{equation}
    \hat{\mathbf{y}}_i^{(\ell)} = \mathrm{softmax}\!\left({\boldsymbol{\theta}^{(\ell)}}^{\!\top} \tilde{\mathbf{h}}_i^{(\ell)} + \mathbf{b}^{(\ell)}\right),
\end{equation}
where $\boldsymbol{\theta}^{(\ell)} \in \mathbb{R}^{d \times C}$ and $\mathbf{b}^{(\ell)} \in \mathbb{R}^{C}$ are the learnable weight and bias parameters of the probe at layer $\ell$.
The probe is optimized by minimizing the soft cross-entropy loss over $n$ training samples:
\begin{equation}
    \mathcal{J}\!\left(\boldsymbol{\theta}^{(\ell)}\right)
    = -\frac{1}{n}\sum_{i=1}^{n}\sum_{c=1}^{C}
      y_{i,c}\log \hat{y}_{i,c}^{(\ell)},
\end{equation}
where $y_{i,c}$ and $\hat{y}_{i,c}^{(\ell)}$ denote the $c$-th element of the soft label $\mathbf{y}_i$ and the predicted distribution $\hat{\mathbf{y}}_i^{(\ell)}$, respectively.

To train and evaluate the linear probes, we construct a tri-modal conflict dataset following the procedure introduced in Section~\ref{sec:Data Construction}.
For each model under evaluation, we independently sample 1{,}000 instances per modality category, yielding a class-balanced pool of 3{,}000 samples in total.
The pool is partitioned into training, validation, and test sets at an 8:1:1 ratio, with class balance strictly maintained within each split.
Each per-layer probe is trained for 200 epochs using the Adam optimizer with a learning rate of 1e-3 and a batch size of 256, and the checkpoint with the lowest validation loss is selected for evaluation.

\begin{figure}[t]
  \centering
  \includegraphics[width=\columnwidth]{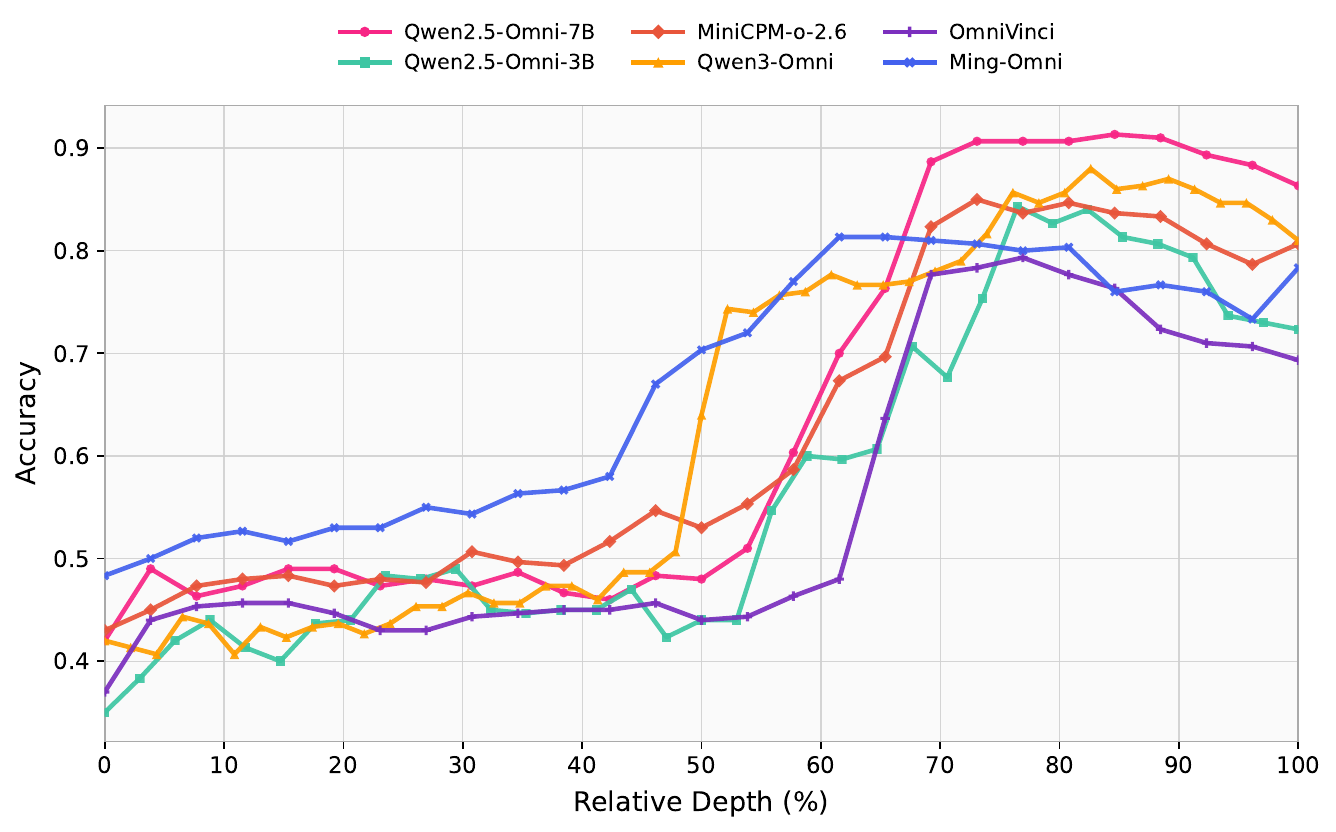}
  \caption{Layer-wise modality preference probe accuracy for all evaluated OLLMs. Qwen3-Omni refers to Qwen3-Omni-30B-A3B-Instruct, and Ming-Omni refers to Ming-Lite-Omni 1.5.}
  \label{fig:probe_curve}
\end{figure}

\subsection{How Preference Emerges}
\textbf{Modality preference is absent in the shallow layers, then emerges in the mid-to-late layers and gradually stabilizes as depth increases.}
Figure~\ref{fig:probe_curve} presents the accuracy curves of modality preference probes across relative layer depth for all evaluated OLLMs. In the first 30\% of layers, probe accuracy for all models remains near chance level, approximately 0.30 to 0.55, indicating that these layers primarily encode low-level features and have not yet developed modality preference signals. Between 40\% and 70\% depth, all models undergo a sharp increase in probe accuracy, with modality preference rising rapidly within this interval. For instance, Qwen2.5-Omni-7B jumps from approximately 0.50 to around 0.90, and MiniCPM-o-2.6 rises from around 0.50 to 0.80, while OmniVinci and Qwen2.5-Omni-3B complete a jump from around 0.45 to above 0.70 within the same interval. Among all OLLMs, Qwen2.5-Omni-7B achieves the highest peak accuracy of approximately 0.90. Beyond 80\% depth, probe accuracy begins to decline to varying degrees across all models. This decline in the final layers aligns with prior findings that the last layers tend to compress intermediate representations into task-specific output distributions, attenuating modality-specific signals that peak at earlier depths~\cite{tenney2019bert,skean2025layer}.

\begin{figure}[t]
  \centering
  \includegraphics[width=\columnwidth]{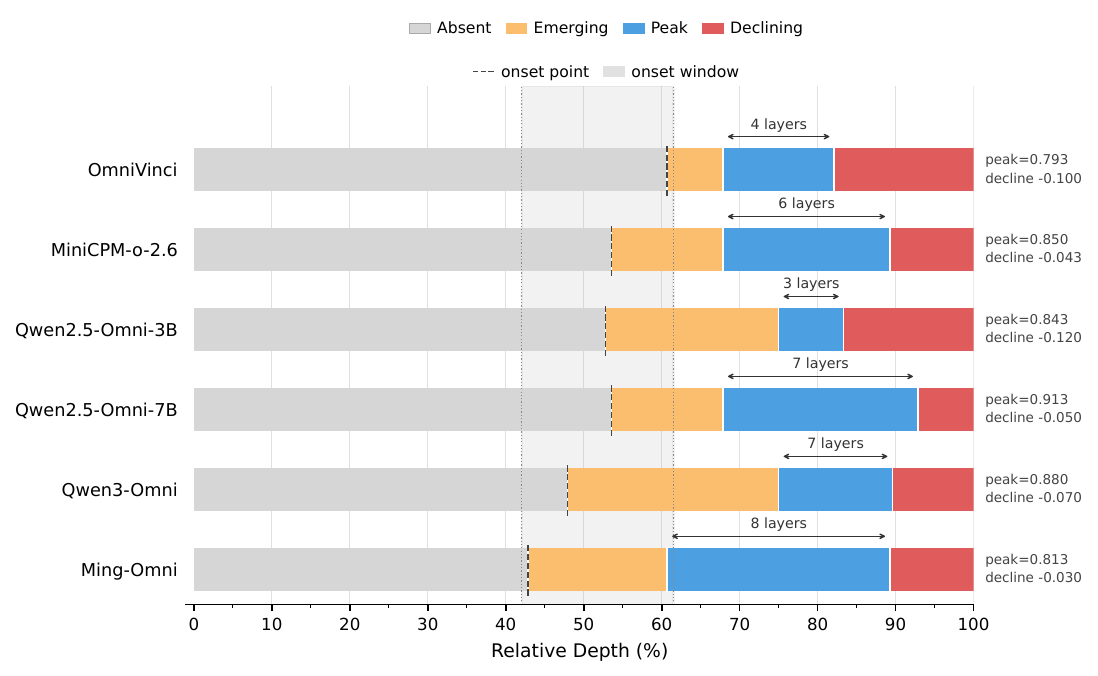}
  \caption{Illustration of the four-phase decomposition of modality preference emergence across evaluated OLLMs along relative layer depth. Qwen3-Omni refers to Qwen3-Omni-30B-A3B-Instruct, and Ming-Omni refers to Ming-Lite-Omni 1.5.}
  \label{fig:probe_gante}
\end{figure}

\textbf{Larger models encode modality preferences earlier with a milder decline, while smaller models encode them later with a more pronounced decline.}
We further characterize the emergence process of modality preference into four phases: \textit{Absent}, \textit{Emerging}, \textit{Peak}, and \textit{Declining}. In the \textit{Absent} phase, the probe accuracy remains low, indicating that modality preference signals have not yet formed. The \textit{Emerging} phase marks a sharp increase in probe accuracy. To identify its starting point, we compute the median of layer-wise accuracy differences over the first 40\% of layers and add three times the median absolute deviation (MAD) as a threshold; the first layer exceeding this threshold is defined as the onset point. Layers where probe accuracy exceeds 95\% of the maximum are assigned to the \textit{Peak} phase, in which modality preference is most pronounced. The \textit{Declining} phase begins when accuracy drops more than 2\% from the peak value with at least two consecutive layers of decrease. 
Figure~\ref{fig:probe_gante} illustrates this four-phase decomposition for all evaluated OLLMs. 
Among all models, Qwen3-Omni-30B-A3B-Instruct and Ming-Lite-Omni 1.5 exhibit the earliest onset points, suggesting that larger-scale models tend to manifest modality preference at shallower relative depths. Qwen2.5-Omni-3B, the smallest model evaluated, sustains the \textit{Peak} phase with an accuracy decline of $-0.120$, whereas Ming-Lite-Omni 1.5 maintains the \textit{Peak} phase with a decline of only $-0.030$, indicating that smaller models tend to exhibit more pronounced preference decline.

\subsection{Representation-Level Analysis}
\begin{figure}[t]
  \centering
  \includegraphics[width=\columnwidth]{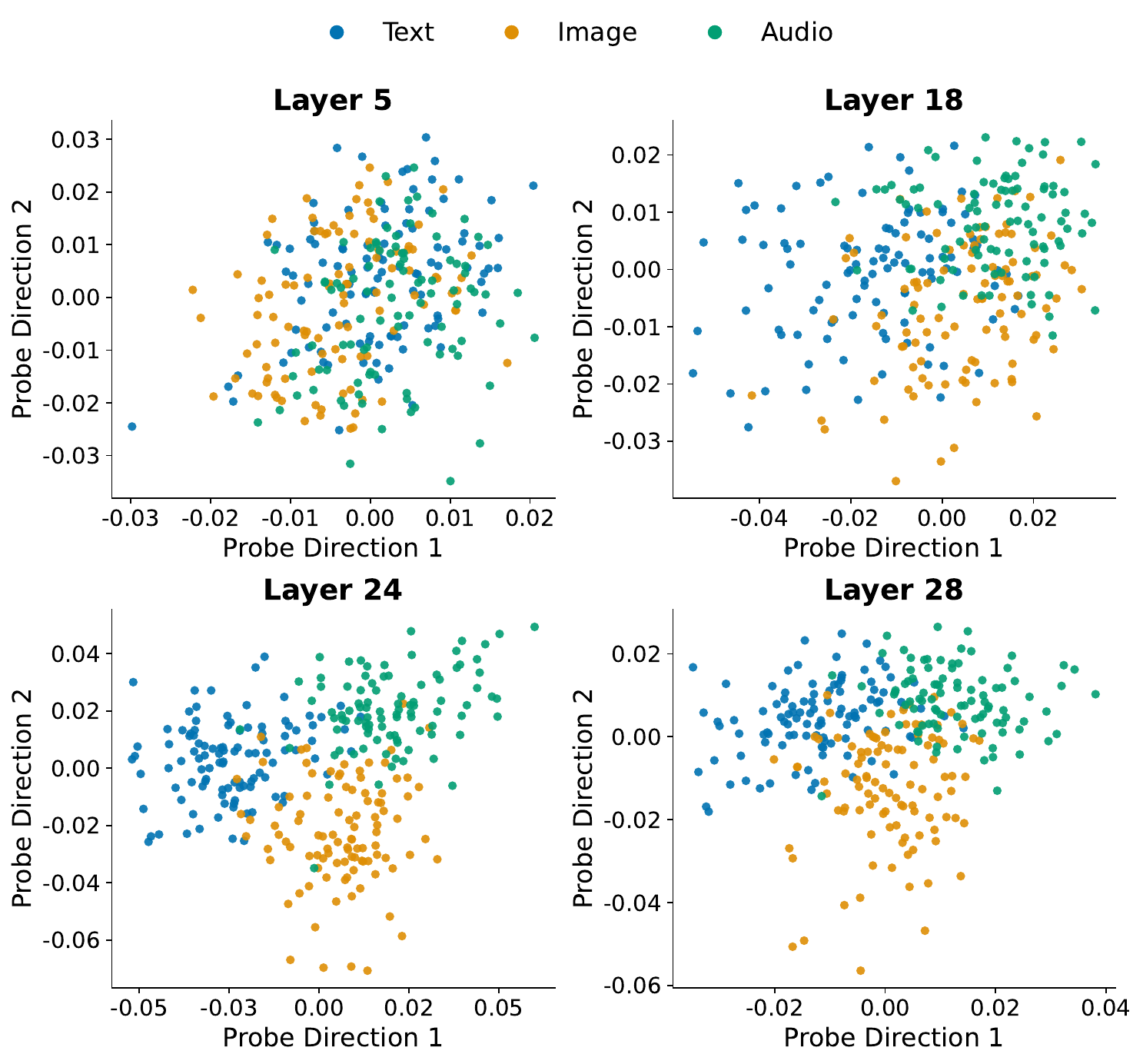}
  \caption{SVD projections of hidden states onto the top two right singular vectors of the probe weight matrix at four representative layers of Qwen2.5-Omni-7B.}
  \label{fig:probe_svd}
\end{figure}

To further examine how modality preference is encoded in hidden states, we perform singular value decomposition on the probe weight matrix $\mathbf{W}^{(\ell)} \in \mathbb{R}^{C \times d}$ at four representative layers of Qwen2.5-Omni-7B and project the hidden states onto the top two right singular vectors. As shown in Figure~\ref{fig:probe_svd}, the projections reveal a layerwise progression: modality preference representations evolve from fully mixed in early layers to well-clustered in mid-to-late layers, and then become partially diffused near the output. Specifically, at Layer~5, samples of all three modality categories are entirely interleaved in the projected space with no discernible cluster structure. By Layer~18, samples with different labels begin to occupy partially distinct regions, though substantial overlap remains. At Layer~24, the separation reaches its clearest form, with the three categories forming distinguishable clusters and minimal inter-class overlap. At Layer~28, cluster boundaries become less defined and inter-class overlap increases compared to Layer~24.

\begin{table}[tb]
\centering
\resizebox{\columnwidth}{!}{%
\begin{tabular}{ccc}
\hline
Dataset & Target Modality & Interfering Modality \\
\hline
POPE & Visual & Text \\
AVHBench (V$\rightarrow$A) & Audio & Visual, Text \\
AVHBench (A$\rightarrow$V) & Visual & Audio, Text \\
AHa-Bench & Audio & Text \\
\hline
\end{tabular}
}
\caption{Target and interfering modality definitions for each hallucination benchmark.}
\label{tab:target_interfere}
\end{table}

\section{Diagnosing Preference-Induced Hallucination}

\finding{Findings 3. Our modality preference probe can also serve as a practical and effective tool for cross-modal hallucination diagnosis.}

This section first analyzes the intrinsic relationship between hallucination occurrence and preference probability, then conducts hallucination detection experiments on benchmark datasets using probes to validate the effectiveness of this approach, and finally presents case studies.

\begin{table}[t]
\centering
\small
\begin{tabular}{cc}
\hline
Dataset & p-value \\
\hline
POPE & 1.08e-60 \\
AVHBench (V$\rightarrow$A) & 4.77e-51 \\
AVHBench (A$\rightarrow$V) & 3.54e-30 \\
AHa-Bench & 1.92e-32 \\
\hline
\end{tabular}
\caption{Mann-Whitney U test p-values across hallucination benchmarks.}
\label{tab:p_value_results}
\vspace{-20pt}
\end{table}

\begin{figure*}[h]
  \centering
  \subfloat[POPE.]{\includegraphics[width=0.5\columnwidth]{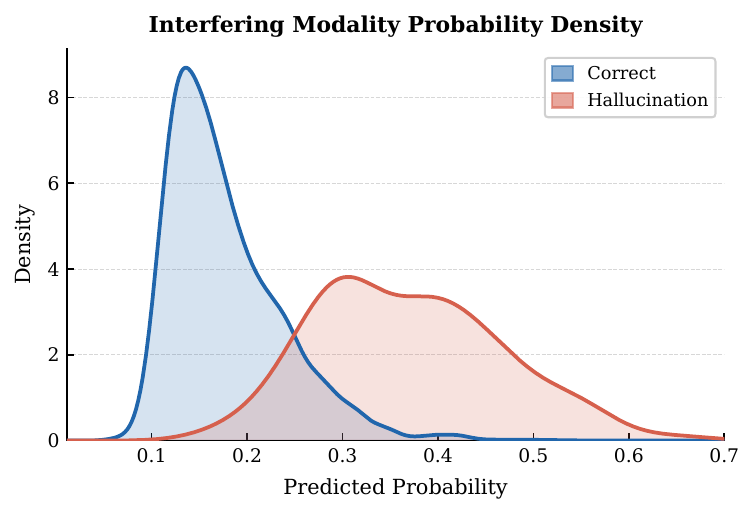}}
  \hfill
  \subfloat[AVHBench(V→A).]{\includegraphics[width=0.5\columnwidth]{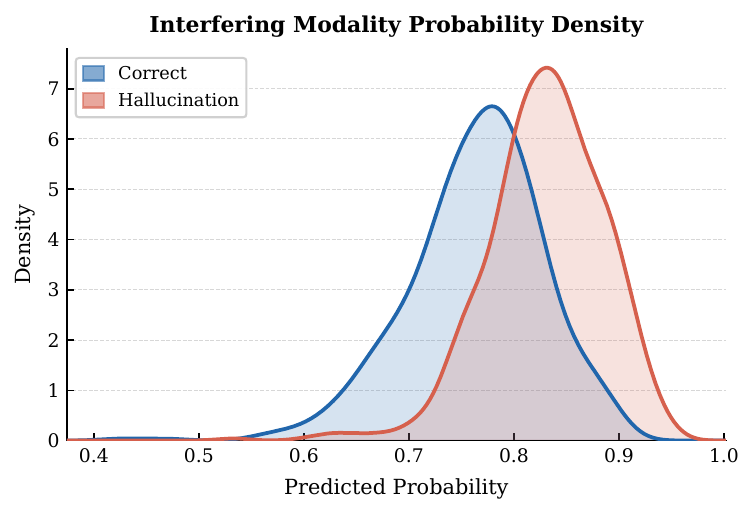}}
  \hfill
  \subfloat[AVHBench(A→V).]{\includegraphics[width=0.5\columnwidth]{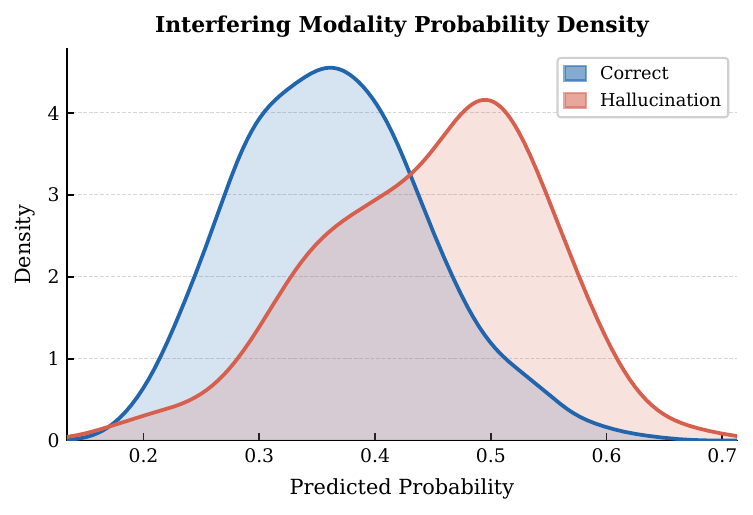}}
  \hfill
  \subfloat[AHa-Bench.]{\includegraphics[width=0.5\columnwidth]{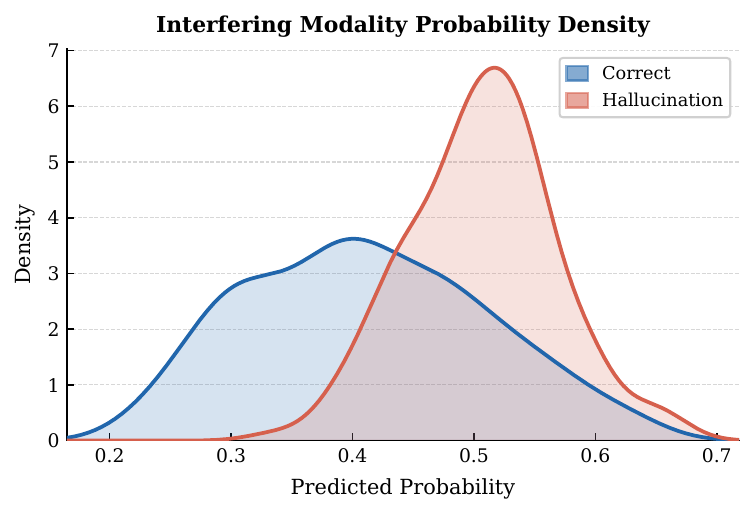}}
  \caption{Density distributions of interfering modality prediction probabilities from layer-wise linear probes on POPE, AVHBench Video-driven Audio, AVHBench Audio-driven Video, and AHa-Bench. The model used is Qwen2.5-Omni-7B.}
  \label{fig:interfering_kde}
\end{figure*}

\begin{table*}[h]
\centering
\arrayrulecolor{black}
\resizebox{\textwidth}{!}{
\begin{tabular}{cccccccccccccc}
\toprule
\multirow{2}{*}{Method} & \multicolumn{3}{c}{POPE} & \multicolumn{3}{c}{AVHBench(V→A)} & \multicolumn{3}{c}{AVHBench(A→V)} & \multicolumn{3}{c}{AHa-Bench} \\
\cmidrule(lr){2-4}\cmidrule(lr){5-7}\cmidrule(lr){8-10}\cmidrule(lr){11-13}
& AUROC & AUPRC & F1-Score & AUROC & AUPRC & F1-Score & AUROC & AUPRC & F1-Score & AUROC & AUPRC & F1-Score \\
\midrule
\multicolumn{13}{c}{Qwen2.5-Omni-7B} \\
\midrule
Random & 0.50 & 0.02 & 0.04 & 0.50 & 0.50 & 0.67 & 0.50 & 0.15 & 0.26 & 0.50 & 0.32 & 0.48 \\
Early Probe & 0.52 & 0.03 & 0.06 & 0.52 & 0.51 & 0.67 & 0.49 & 0.14 & 0.26 & 0.49 & 0.30 & 0.49 \\
\rowcolor[rgb]{0.941,0.941,1} Ours & \textbf{0.96} & \textbf{0.51} & \textbf{0.54} & \textbf{0.76} & \textbf{0.75}  & \textbf{0.72} & \textbf{0.75} & \textbf{0.53} & \textbf{0.52} & \textbf{0.84} & \textbf{0.72} & \textbf{0.69} \\
\midrule
\multicolumn{13}{c}{MiniCPM-o-2.6} \\
\midrule
Random & 0.50 & 0.05 & 0.10 & 0.50 & 0.61 & 0.76 & 0.50 & 0.31 & 0.47 & 0.50 & 0.37 & 0.54 \\
Early Probe & 0.49 & 0.05 & 0.09 & 0.51 & 0.39 & 0.55 & 0.49 & 0.30 & 0.48 & 0.49 & 0.36 & 0.54 \\
\rowcolor[rgb]{0.941,0.941,1} Ours & \textbf{0.99} & \textbf{0.83} & \textbf{0.75} & \textbf{0.89} & \textbf{0.82} & \textbf{0.76} & \textbf{0.76} & \textbf{0.65} & \textbf{0.67} & \textbf{0.75} & \textbf{0.65} & \textbf{0.64} \\
\midrule
\multicolumn{13}{c}{Qwen3-Omni-30B-A3B-Instruct} \\
\midrule
Random & 0.50 & 0.08 & 0.15 & 0.50 & 0.40 & 0.57 & 0.50 & 0.24 & 0.39 & 0.50 & 0.23 & 0.37 \\
Early Probe & 0.51 & 0.09 & 0.15 & 0.51 & 0.42 & 0.58 & 0.51 & 0.25 & 0.40 & 0.51 & 0.28 & 0.38 \\
\rowcolor[rgb]{0.941,0.941,1} Ours & \textbf{0.87} & \textbf{0.53} & \textbf{0.55} & \textbf{0.80} & \textbf{0.75} & \textbf{0.62} & \textbf{0.72} & \textbf{0.74} & \textbf{0.67} & \textbf{0.80} & \textbf{0.70} & \textbf{0.67} \\
\bottomrule
\end{tabular}
}
\caption{Hallucination detection performance of our probe-based method against two baselines across models and benchmarks.}
\label{tab:hallucination_detection}
\end{table*}

\begin{figure*}[t]
  \centering
  \includegraphics[width=\textwidth]{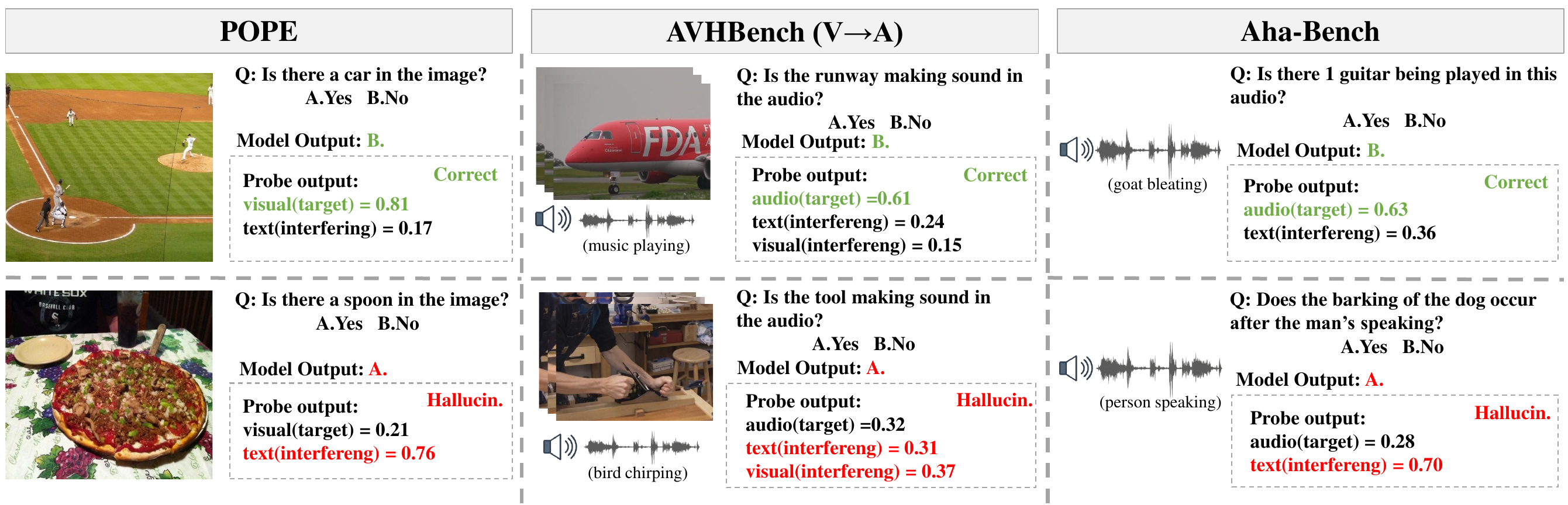}
  \caption{Representative cases of the linear probe detecting hallucinations by predicting the interfering modality preference probability.}
  \label{fig:case_study}
\end{figure*}

\subsection{Modality Preference Causes Cross-modal Hallucination}
The preceding analysis has shown that linear probes can capture modality preference signals from the internal representations of OLLMs.
A natural follow-up question is whether these internal preferences affect model reliability on downstream tasks. To investigate this question, we conduct experiments on three hallucination benchmarks spanning diverse modality combinations.
Specifically, we select POPE~\cite{li2023evaluating}, AVHBench~\cite{sung2024avhbench} (with two sub-tasks: Audio-driven Video Hallucination and Video-driven Audio Hallucination), and AHa-Bench~\cite{kuan2025can} for hallucination detection.
To investigate the relationship between modality preference and hallucination, we define two roles for each benchmark: the \textit{target modality}, which the model should attend to for correct answering, and the \textit{interfering modality}, which may mislead the model. The complete definitions for all benchmarks are provided in Table~\ref{tab:target_interfere}.

\textbf{The occurrence of hallucinations is consistently accompanied by an abnormal increase in the predicted preference probability for the interfering modality.}
Using the linear probe from the layer with the highest preference prediction accuracy on Qwen2.5-Omni-7B, we compute the predicted probability of the interfering modality for each sample and plot the probability density distributions for correct and hallucinated samples separately. 
As shown in Figure~\ref{fig:interfering_kde}, notably, across all benchmarks, the interfering modality probability distribution of hallucinated samples is significantly shifted toward higher values compared to correct samples.
To quantify this relationship, we conduct the Mann-Whitney U test to assess the distributional difference in interfering modality probability between correct and hallucinated samples. As shown in Table~\ref{tab:p_value_results}, the p-values across all four datasets are extremely low, with POPE at 1.08e-60, the two sub-tasks of AVHBench at 4.77e-51 and 3.54e-30, and AHa-Bench at 1.92e-32. This indicates that the preference probability distribution of the interfering modality for hallucinated samples differs significantly from that of correct samples. Consequently, the interfering modality probability predicted by a linear probe can serve as a reliable indicator for detecting hallucinations.

\subsection{Probe-based Diagnosis}
We employ the linear probe from the layer exhibiting the strongest modality preference as a hallucination detector across all four hallucination detection tasks. For each sample, the probe computes the predicted probability assigned to the interfering modality, which serves as the hallucination risk score.
To evaluate the probe's effectiveness for hallucination detection, we restrict our evaluation set to samples where the correct answer is "no." In this setting, a model response of "yes" indicates the occurrence of a hallucination.  
To mitigate the known affirmative bias exhibited by LLMs ~\cite{tjuatja2024llms,li2023evaluating}, we reformulate all yes/no questions into a binary multiple-choice format. 
To further avoid potential position bias ~\cite{zheng2023large}, we randomize the order of options across samples, ensuring that "Yes" and "No" are equally likely to appear as option A or B.
We compare against two baselines. The first is a Random baseline, which serves as a chance-level reference to verify whether our probe captures a meaningful signal related to hallucination. The second is an Early Probe, which uses the probe from Layer 1 as the detector, designed to examine whether the detection signal is layer-specific or already present in early layers before modality preference emerges.
We evaluate all methods using three complementary metrics: AUROC for threshold-free assessment of the probe's ability to distinguish hallucinated from non-hallucinated samples, AUPRC for evaluating detection reliability under class imbalance, and Optimal F1 for measuring the best achievable balance between precision and recall across all decision thresholds.

\textbf{Our probe can serve as a diagnostic tool for detecting hallucination phenomena arising from modality preference.}
As shown in Table~\ref{tab:hallucination_detection}, across all models and benchmarks, the Early Probe yields performance nearly indistinguishable from the Random baseline, with AUROC values hovering around 0.50 in almost all settings, indicating that the hallucination detection signal has not yet emerged in the early layers of the model. 
In contrast, our probe achieves remarkable detection performance across all settings, with an average AUROC of 0.94 on POPE across the three models.
Specifically, our method attains AUROC scores of 0.96, 0.99, and 0.87 on Qwen2.5-Omni-7B, MiniCPM-o-2.6, and Qwen3-Omni-30B-A3B-Instruct, respectively, with corresponding AUPRC scores of 0.51, 0.83, and 0.53, and Optimal F1 scores of 0.54, 0.75, and 0.55.
For AVHBench(V→A) and AVHBench(A→V), the AUROC consistently exceeds 0.72 across all models. Notably, MiniCPM-o-2.6 achieves an AUROC of 0.89 and an AUPRC of 0.82 on the Video-driven Audio sub-task, while Qwen3-Omni-30B-A3B-Instruct reaches an AUPRC of 0.74 and an Optimal F1 of 0.67 on the Audio-driven Video sub-task.
On AHa-Bench, our method achieves AUROC scores ranging from 0.75 to 0.84, with AUPRC values between 0.65 and 0.72 and Optimal F1 scores between 0.64 and 0.69, confirming the effectiveness of the probe-based approach across diverse hallucination scenarios.

\subsection{Case Study}
Figure~\ref{fig:case_study} presents representative cases from three benchmarks, illustrating how our probe diagnoses hallucinations through the interfering modality preference probability.
On POPE, the visual (target) preference probability reaches 0.81 when the model answers correctly, but drops sharply to 0.21 when hallucination occurs, while the text (interfering) preference probability surges to 0.76. On AVHBench (V→A), the audio (target) preference probability decreases from 0.61 to 0.32 during hallucination, with the combined probability of the two interfering modalities exceeding that of the target modality. On AHa-Bench, the audio (target) preference probability falls to 0.28 in the hallucination case, while the text (interfering) preference probability rises as high as 0.70.

\section{Conclusion}
In this paper, we systematically investigate modality preference in OLLMs across three dimensions: behavioral evaluation, mechanistic analysis, and hallucination detection. Our tri-modal conflict framework reveals that OLLMs predominantly favor vision while systematically neglecting audio. Layer-wise probing further shows that modality preference emerges progressively in mid-to-late layers. Finally, we demonstrate that hallucinations correlate with abnormal preference shifts, which can be effectively detected via linear probes.

\bibliography{custom}

\begin{thebibliography}{47}
\providecommand{\natexlab}[1]{#1}

\bibitem[{AI et~al.(2025)AI, Gong, Zou, Zheng, Zhou, Yan, Jin, Shen, Zheng, Wang et~al.}]{ai2025ming}
Inclusion AI, Biao Gong, Cheng Zou, Chuanyang Zheng, Chunluan Zhou, Canxiang Yan, Chunxiang Jin, Chunjie Shen, Dandan Zheng, Fudong Wang, and 1 others. 2025.
\newblock Ming-omni: A unified multimodal model for perception and generation.
\newblock \emph{arXiv preprint arXiv:2506.09344}.

\bibitem[{Alain and Bengio(2016)}]{alain2016understanding}
Guillaume Alain and Yoshua Bengio. 2016.
\newblock Understanding intermediate layers using linear classifier probes.
\newblock \emph{arXiv preprint arXiv:1610.01644}.

\bibitem[{Bai et~al.(2023)Bai, Bai, Yang, Wang, Tan, Wang, Lin, Zhou, and Zhou}]{bai2023qwenvlversatilevisionlanguagemodel}
Jinze Bai, Shuai Bai, Shusheng Yang, Shijie Wang, Sinan Tan, Peng Wang, Junyang Lin, Chang Zhou, and Jingren Zhou. 2023.
\newblock Qwen-vl: A versatile vision-language model for understanding, localization, text reading, and beyond.
\newblock \emph{arXiv preprint arXiv:2308.12966}.

\bibitem[{Bai et~al.(2024)Bai, Wang, Xiao, He, Han, Zhang, and Shou}]{bai2024hallucination}
Zechen Bai, Pichao Wang, Tianjun Xiao, Tong He, Zongbo Han, Zheng Zhang, and Mike~Zheng Shou. 2024.
\newblock Hallucination of multimodal large language models: A survey.
\newblock \emph{arXiv preprint arXiv:2404.18930}.

\bibitem[{Belinkov(2022)}]{belinkov2022probing}
Yonatan Belinkov. 2022.
\newblock Probing classifiers: Promises, shortcomings, and advances.
\newblock \emph{Computational Linguistics}, 48(1):207--219.

\bibitem[{Chen et~al.(2024{\natexlab{a}})Chen, Cao, Zhang, and Lu}]{chen2024quantifying}
Meiqi Chen, Yixin Cao, Yan Zhang, and Chaochao Lu. 2024{\natexlab{a}}.
\newblock Quantifying and mitigating unimodal biases in multimodal large language models: A causal perspective.
\newblock In \emph{Findings of the Association for Computational Linguistics: EMNLP 2024}, pages 16449--16469.

\bibitem[{Chen et~al.(2024{\natexlab{b}})Chen, Wu, Wang, Su, Chen, Xing, Zhong, Zhang, Zhu, Lu et~al.}]{chen2024internvl}
Zhe Chen, Jiannan Wu, Wenhai Wang, Weijie Su, Guo Chen, Sen Xing, Muyan Zhong, Qinglong Zhang, Xizhou Zhu, Lewei Lu, and 1 others. 2024{\natexlab{b}}.
\newblock Internvl: Scaling up vision foundation models and aligning for generic visual-linguistic tasks.
\newblock In \emph{Proceedings of the IEEE/CVF conference on computer vision and pattern recognition}, pages 24185--24198.

\bibitem[{Comanici et~al.(2025)Comanici, Bieber, Schaekermann, Pasupat, Sachdeva, Dhillon, Blistein, Ram, Zhang, Rosen et~al.}]{comanici2025gemini}
Gheorghe Comanici, Eric Bieber, Mike Schaekermann, Ice Pasupat, Noveen Sachdeva, Inderjit Dhillon, Marcel Blistein, Ori Ram, Dan Zhang, Evan Rosen, and 1 others. 2025.
\newblock Gemini 2.5: Pushing the frontier with advanced reasoning, multimodality, long context, and next generation agentic capabilities.
\newblock \emph{arXiv preprint arXiv:2507.06261}.

\bibitem[{Deng et~al.(2025)Deng, Cao, Chen, and Hooi}]{deng2025words}
Ailin Deng, Tri Cao, Zhirui Chen, and Bryan Hooi. 2025.
\newblock Words or vision: Do vision-language models have blind faith in text?
\newblock In \emph{Proceedings of the Computer Vision and Pattern Recognition Conference}, pages 3867--3876.

\bibitem[{Duki{\'c} and {\v{S}}najder(2024)}]{dukic2024looking}
David Duki{\'c} and Jan {\v{S}}najder. 2024.
\newblock Looking right is sometimes right: Investigating the capabilities of decoder-only llms for sequence labeling.
\newblock In \emph{Findings of the Association for Computational Linguistics: ACL 2024}, pages 14168--14181.

\bibitem[{Fu et~al.(2024)Fu, Lin, Long, Shen, Dai, Zhao, Zhang, Dong, Li, Wang et~al.}]{fu2024vita}
Chaoyou Fu, Haojia Lin, Zuwei Long, Yunhang Shen, Yuhang Dai, Meng Zhao, Yi-Fan Zhang, Shaoqi Dong, Yangze Li, Xiong Wang, and 1 others. 2024.
\newblock Vita: Towards open-source interactive omni multimodal llm.
\newblock \emph{arXiv preprint arXiv:2408.05211}.

\bibitem[{Hinton et~al.(2015)Hinton, Vinyals, and Dean}]{hinton2015distilling}
Geoffrey Hinton, Oriol Vinyals, and Jeff Dean. 2015.
\newblock Distilling the knowledge in a neural network.
\newblock \emph{arXiv preprint arXiv:1503.02531}.

\bibitem[{Hua et~al.(2025)Hua, Yun, and Pavlick}]{hua2025vision}
Tianze Hua, Tian Yun, and Ellie Pavlick. 2025.
\newblock How do vision-language models process conflicting information across modalities?
\newblock \emph{arXiv preprint arXiv:2507.01790}.

\bibitem[{Hurst et~al.(2024)Hurst, Lerer, Goucher, Perelman, Ramesh, Clark, Ostrow, Welihinda, Hayes, Radford et~al.}]{hurst2024gpt}
Aaron Hurst, Adam Lerer, Adam~P Goucher, Adam Perelman, Aditya Ramesh, Aidan Clark, AJ~Ostrow, Akila Welihinda, Alan Hayes, Alec Radford, and 1 others. 2024.
\newblock Gpt-4o system card.
\newblock \emph{arXiv preprint arXiv:2410.21276}.

\bibitem[{Jiang et~al.(2025)Jiang, Liang, Wang, Dong, Chang, Yu, Du, Liu, and Qin}]{jiang2025specific}
Shixin Jiang, Jiafeng Liang, Jiyuan Wang, Xuan Dong, Heng Chang, Weijiang Yu, Jinhua Du, Ming Liu, and Bing Qin. 2025.
\newblock From specific-mllms to omni-mllms: a survey on mllms aligned with multi-modalities.
\newblock In \emph{Findings of the Association for Computational Linguistics: ACL 2025}, pages 8617--8652.

\bibitem[{Kornblith et~al.(2019)Kornblith, Norouzi, Lee, and Hinton}]{kornblith2019similarity}
Simon Kornblith, Mohammad Norouzi, Honglak Lee, and Geoffrey Hinton. 2019.
\newblock Similarity of neural network representations revisited.
\newblock In \emph{International conference on machine learning}, pages 3519--3529. PMlR.

\bibitem[{Kuan and Lee(2025)}]{kuan2025can}
Chun-Yi Kuan and Hung-yi Lee. 2025.
\newblock Can large audio-language models truly hear? tackling hallucinations with multi-task assessment and stepwise audio reasoning.
\newblock In \emph{ICASSP 2025-2025 IEEE International Conference on Acoustics, Speech and Signal Processing (ICASSP)}, pages 1--5. IEEE.

\bibitem[{Kwon et~al.(2025)Kwon, Kim, Lee, Choi, and Kim}]{kwon2025see}
JuneHyoung Kwon, MiHyeon Kim, Eunju Lee, Juhwan Choi, and YoungBin Kim. 2025.
\newblock See-saw modality balance: See gradient, and sew impaired vision-language balance to mitigate dominant modality bias.
\newblock In \emph{Proceedings of the 2025 Conference of the Nations of the Americas Chapter of the Association for Computational Linguistics: Human Language Technologies (Volume 1: Long Papers)}, pages 4364--4378.

\bibitem[{Leng et~al.(2024)Leng, Zhang, Chen, Li, Lu, Miao, and Bing}]{leng2024mitigating}
Sicong Leng, Hang Zhang, Guanzheng Chen, Xin Li, Shijian Lu, Chunyan Miao, and Lidong Bing. 2024.
\newblock Mitigating object hallucinations in large vision-language models through visual contrastive decoding.
\newblock In \emph{Proceedings of the IEEE/CVF Conference on Computer Vision and Pattern Recognition}, pages 13872--13882.

\bibitem[{Li et~al.(2023{\natexlab{a}})Li, Li, Savarese, and Hoi}]{li2023blip}
Junnan Li, Dongxu Li, Silvio Savarese, and Steven Hoi. 2023{\natexlab{a}}.
\newblock Blip-2: Bootstrapping language-image pre-training with frozen image encoders and large language models.
\newblock In \emph{International conference on machine learning}, pages 19730--19742. PMLR.

\bibitem[{Li et~al.(2025)Li, Liu, Zhang, Chen, Li, Li, Liu, Ming, Dong, Pan et~al.}]{li2025baichuan}
Yadong Li, Jun Liu, Tao Zhang, Song Chen, Tianpeng Li, Zehuan Li, Lijun Liu, Lingfeng Ming, Guosheng Dong, Da~Pan, and 1 others. 2025.
\newblock Baichuan-omni-1.5 technical report.
\newblock \emph{arXiv preprint arXiv:2501.15368}.

\bibitem[{Li et~al.(2023{\natexlab{b}})Li, Du, Zhou, Wang, Zhao, and Wen}]{li2023evaluating}
Yifan Li, Yifan Du, Kun Zhou, Jinpeng Wang, Wayne~Xin Zhao, and Ji-Rong Wen. 2023{\natexlab{b}}.
\newblock Evaluating object hallucination in large vision-language models.
\newblock In \emph{Proceedings of the 2023 conference on empirical methods in natural language processing}, pages 292--305.

\bibitem[{Li et~al.(2024)Li, Ma, Zhang, Yuan, Zhu, Guo, Liang, Liu, Wang, Yang et~al.}]{li2024omnibench}
Yizhi Li, Yinghao Ma, Ge~Zhang, Ruibin Yuan, Kang Zhu, Hangyu Guo, Yiming Liang, Jiaheng Liu, Zekun Wang, Jian Yang, and 1 others. 2024.
\newblock Omnibench: Towards the future of universal omni-language models.
\newblock \emph{arXiv preprint arXiv:2409.15272}.

\bibitem[{Lin et~al.(2025)Lin, Basu, Beigi, Manjunatha, Rossi, Wang, Zhou, Balasubramanian, Zarei, Rezaei et~al.}]{lin2025survey}
Zihao Lin, Samyadeep Basu, Mohammad Beigi, Varun Manjunatha, Ryan~A Rossi, Zichao Wang, Yufan Zhou, Sriram Balasubramanian, Arman Zarei, Keivan Rezaei, and 1 others. 2025.
\newblock A survey on mechanistic interpretability for multi-modal foundation models.
\newblock \emph{arXiv preprint arXiv:2502.17516}.

\bibitem[{Liu et~al.(2024)Liu, Li, Li, and Lee}]{liu2024improved}
Haotian Liu, Chunyuan Li, Yuheng Li, and Yong~Jae Lee. 2024.
\newblock Improved baselines with visual instruction tuning.
\newblock In \emph{Proceedings of the IEEE/CVF conference on computer vision and pattern recognition}, pages 26296--26306.

\bibitem[{Liu et~al.(2023)Liu, Li, Wu, and Lee}]{liu2023visual}
Haotian Liu, Chunyuan Li, Qingyang Wu, and Yong~Jae Lee. 2023.
\newblock Visual instruction tuning.
\newblock \emph{Advances in neural information processing systems}, 36:34892--34916.

\bibitem[{Lou et~al.(2025)Lou, Li, Ji, and Yang}]{lou2025sae}
Hantao Lou, Changye Li, Jiaming Ji, and Yaodong Yang. 2025.
\newblock Sae-v: Interpreting multimodal models for enhanced alignment.
\newblock \emph{arXiv preprint arXiv:2502.17514}.

\bibitem[{Meng et~al.(2022)Meng, Bau, Andonian, and Belinkov}]{meng2022locating}
Kevin Meng, David Bau, Alex Andonian, and Yonatan Belinkov. 2022.
\newblock Locating and editing factual associations in gpt.
\newblock \emph{Advances in neural information processing systems}, 35:17359--17372.

\bibitem[{Neelakantan et~al.(2022)Neelakantan, Xu, Puri, Radford, Han, Tworek, Yuan, Tezak, Kim, Hallacy et~al.}]{neelakantan2022text}
Arvind Neelakantan, Tao Xu, Raul Puri, Alec Radford, Jesse~Michael Han, Jerry Tworek, Qiming Yuan, Nikolas Tezak, Jong~Wook Kim, Chris Hallacy, and 1 others. 2022.
\newblock Text and code embeddings by contrastive pre-training.
\newblock \emph{arXiv preprint arXiv:2201.10005}.

\bibitem[{Parcalabescu et~al.(2022)Parcalabescu, Cafagna, Muradjan, Frank, Calixto, and Gatt}]{parcalabescu2022valse}
Letitia Parcalabescu, Michele Cafagna, Lilitta Muradjan, Anette Frank, Iacer Calixto, and Albert Gatt. 2022.
\newblock Valse: A task-independent benchmark for vision and language models centered on linguistic phenomena.
\newblock In \emph{Proceedings of the 60th Annual Meeting of the Association for Computational Linguistics (Volume 1: Long Papers)}, pages 8253--8280.

\bibitem[{Petroni et~al.(2019)Petroni, Rockt{\"a}schel, Riedel, Lewis, Bakhtin, Wu, and Miller}]{petroni2019language}
Fabio Petroni, Tim Rockt{\"a}schel, Sebastian Riedel, Patrick Lewis, Anton Bakhtin, Yuxiang Wu, and Alexander Miller. 2019.
\newblock Language models as knowledge bases?
\newblock In \emph{Proceedings of the 2019 conference on empirical methods in natural language processing and the 9th international joint conference on natural language processing (EMNLP-IJCNLP)}, pages 2463--2473.

\bibitem[{Pezeshkpour et~al.(2025)Pezeshkpour, Aminnaseri, and Hruschka}]{pezeshkpour2025mixed}
Pouya Pezeshkpour, Moin Aminnaseri, and Estevam Hruschka. 2025.
\newblock Mixed signals: Decoding vlms’ reasoning and underlying bias in vision-language conflict.
\newblock \emph{arXiv preprint arXiv:2504.08974}.

\bibitem[{Singh et~al.(2025)Singh, Fry, Perelman, Tart, Ganesh, El-Kishky, McLaughlin, Low, Ostrow, Ananthram et~al.}]{singh2025openai}
Aaditya Singh, Adam Fry, Adam Perelman, Adam Tart, Adi Ganesh, Ahmed El-Kishky, Aidan McLaughlin, Aiden Low, AJ~Ostrow, Akhila Ananthram, and 1 others. 2025.
\newblock Openai gpt-5 system card.
\newblock \emph{arXiv preprint arXiv:2601.03267}.

\bibitem[{Skean et~al.(2025)Skean, Arefin, Zhao, Patel, Naghiyev, LeCun, and Shwartz-Ziv}]{skean2025layer}
Oscar Skean, Md~Rifat Arefin, Dan Zhao, Niket Patel, Jalal Naghiyev, Yann LeCun, and Ravid Shwartz-Ziv. 2025.
\newblock Layer by layer: Uncovering hidden representations in language models.
\newblock \emph{arXiv preprint arXiv:2502.02013}.

\bibitem[{Sung-Bin et~al.(2024)Sung-Bin, Hyun-Bin, Lee, Senocak, Chung, and Oh}]{sung2024avhbench}
Kim Sung-Bin, Oh~Hyun-Bin, JungMok Lee, Arda Senocak, Joon~Son Chung, and Tae-Hyun Oh. 2024.
\newblock Avhbench: A cross-modal hallucination benchmark for audio-visual large language models.
\newblock \emph{arXiv preprint arXiv:2410.18325}.

\bibitem[{Team et~al.(2023)Team, Anil, Borgeaud, Alayrac, Yu, Soricut, Schalkwyk, Dai, Hauth, Millican et~al.}]{team2023gemini}
Gemini Team, Rohan Anil, Sebastian Borgeaud, Jean-Baptiste Alayrac, Jiahui Yu, Radu Soricut, Johan Schalkwyk, Andrew~M Dai, Anja Hauth, Katie Millican, and 1 others. 2023.
\newblock Gemini: a family of highly capable multimodal models.
\newblock \emph{arXiv preprint arXiv:2312.11805}.

\bibitem[{Tenney et~al.(2019)Tenney, Das, and Pavlick}]{tenney2019bert}
Ian Tenney, Dipanjan Das, and Ellie Pavlick. 2019.
\newblock Bert rediscovers the classical nlp pipeline.
\newblock In \emph{Proceedings of the 57th annual meeting of the association for computational linguistics}, pages 4593--4601.

\bibitem[{Tjuatja et~al.(2024)Tjuatja, Chen, Wu, Talwalkwar, and Neubig}]{tjuatja2024llms}
Lindia Tjuatja, Valerie Chen, Tongshuang Wu, Ameet Talwalkwar, and Graham Neubig. 2024.
\newblock Do llms exhibit human-like response biases? a case study in survey design.
\newblock \emph{Transactions of the Association for Computational Linguistics}, 12:1011--1026.

\bibitem[{Wang et~al.(2025)Wang, Liu, Huang, Yu, Wang, Sun, Wu, Yuille, Barsoum, and Liu}]{wang2025xmodbench}
Xingrui Wang, Jiang Liu, Chao Huang, Xiaodong Yu, Ze~Wang, Ximeng Sun, Jialian Wu, Alan Yuille, Emad Barsoum, and Zicheng Liu. 2025.
\newblock Xmodbench: Benchmarking cross-modal capabilities and consistency in omni-language models.
\newblock \emph{arXiv preprint arXiv:2510.15148}.

\bibitem[{Xu et~al.(2025{\natexlab{a}})Xu, Guo, He, Hu, He, Bai, Chen, Wang, Fan, Dang, Zhang, Wang, Chu, and Lin}]{xu2025qwen25omnitechnicalreport}
Jin Xu, Zhifang Guo, Jinzheng He, Hangrui Hu, Ting He, Shuai Bai, Keqin Chen, Jialin Wang, Yang Fan, Kai Dang, Bin Zhang, Xiong Wang, Yunfei Chu, and Junyang Lin. 2025{\natexlab{a}}.
\newblock Qwen2.5-omni technical report.
\newblock \emph{arXiv preprint arXiv:2503.20215}.

\bibitem[{Xu et~al.(2025{\natexlab{b}})Xu, Guo, Hu, Chu, Wang, He, Wang, Shi, He, Zhu et~al.}]{xu2025qwen3}
Jin Xu, Zhifang Guo, Hangrui Hu, Yunfei Chu, Xiong Wang, Jinzheng He, Yuxuan Wang, Xian Shi, Ting He, Xinfa Zhu, and 1 others. 2025{\natexlab{b}}.
\newblock Qwen3-omni technical report.
\newblock \emph{arXiv preprint arXiv:2509.17765}.

\bibitem[{Yao et~al.(2024)Yao, Yu, Zhang, Wang, Cui, Zhu, Cai, Li, Zhao, He et~al.}]{yao2024minicpm}
Yuan Yao, Tianyu Yu, Ao~Zhang, Chongyi Wang, Junbo Cui, Hongji Zhu, Tianchi Cai, Haoyu Li, Weilin Zhao, Zhihui He, and 1 others. 2024.
\newblock Minicpm-v: A gpt-4v level mllm on your phone.
\newblock \emph{arXiv preprint arXiv:2408.01800}.

\bibitem[{Ye et~al.(2025)Ye, Yang, Goel, Huang, Zhu, Su, Lin, Cheng, Wan, Tian et~al.}]{ye2025omnivinci}
Hanrong Ye, Chao-Han~Huck Yang, Arushi Goel, Wei Huang, Ligeng Zhu, Yuanhang Su, Sean Lin, An-Chieh Cheng, Zhen Wan, Jinchuan Tian, and 1 others. 2025.
\newblock Omnivinci: Enhancing architecture and data for omni-modal understanding llm.
\newblock \emph{arXiv preprint arXiv:2510.15870}.

\bibitem[{Zhang et~al.(2025)Zhang, Wang, Gong, Shi, Wang, Wang, and Hu}]{zhang2025modalities}
Zhuoran Zhang, Tengyue Wang, Xilin Gong, Yang Shi, Haotian Wang, Di~Wang, and Lijie Hu. 2025.
\newblock When modalities conflict: How unimodal reasoning uncertainty governs preference dynamics in mllms.
\newblock \emph{arXiv preprint arXiv:2511.02243}.

\bibitem[{Zheng et~al.(2023)Zheng, Zhou, Meng, Zhou, and Huang}]{zheng2023large}
Chujie Zheng, Hao Zhou, Fandong Meng, Jie Zhou, and Minlie Huang. 2023.
\newblock Large language models are not robust multiple choice selectors.
\newblock \emph{arXiv preprint arXiv:2309.03882}.

\bibitem[{Zheng et~al.(2025{\natexlab{a}})Zheng, Wu, Wang, and Jiang}]{zheng2025unveiling}
Xinhan Zheng, Huyu Wu, Xueting Wang, and Haiyun Jiang. 2025{\natexlab{a}}.
\newblock Unveiling intrinsic text bias in multimodal large language models through attention key-space analysis.
\newblock \emph{arXiv preprint arXiv:2510.26721}.

\bibitem[{Zheng et~al.(2025{\natexlab{b}})Zheng, Liao, Fu, Lei, Lyu, Jiang, Ren, Chen, Wang, Li et~al.}]{zheng2025mllms}
Xu~Zheng, Chenfei Liao, Yuqian Fu, Kaiyu Lei, Yuanhuiyi Lyu, Lutao Jiang, Bin Ren, Jialei Chen, Jiawen Wang, Chengxin Li, and 1 others. 2025{\natexlab{b}}.
\newblock Mllms are deeply affected by modality bias.
\newblock \emph{arXiv preprint arXiv:2505.18657}.

\end{thebibliography}

\end{document}